\documentclass[twoside,11pt]{article}

\usepackage[utf8]{inputenc}
\usepackage{amsmath}
\usepackage{pgfplots}
\usepackage{wrapfig}
\usepackage[hang]{caption}
\usepackage[colorlinks=true,linkcolor=black,citecolor=black]{hyperref}
\RequirePackage{amssymb}
\RequirePackage{natbib}
\RequirePackage{graphicx}
\bibpunct{(}{)}{;}{a}{,}{,}

\newcommand{\figref}[1]{Figure~\ref{#1}}

\DeclareMathOperator*{\plim}{plim}
\long\def\acks#1{\vskip 0.3in\noindent{\large\bf Acknowledgments}\vskip 0.2in \noindent #1}

\hypersetup{urlcolor=black}

\usetikzlibrary{arrows,positioning,patterns}
\tikzset{hunit/.style={circle,draw,minimum size=0.8cm}}
\tikzset{vunit/.style={circle,draw,minimum size=0.8cm,pattern=crosshatch}}

\definecolor{line1}{rgb}{0.0, 0.0, 1.0}
\definecolor{line2}{rgb}{0.9, 0.7, 0.0}
\definecolor{line3}{rgb}{1.0, 0.0, 0.0}
\definecolor{line4}{rgb}{0.3, 0.8, 0.0}
\definecolor{line5}{rgb}{0.2, 0.2, 0.2}
\definecolor{line6}{rgb}{0.5, 0.7, 1.0}

\begin{document}
	\title{In All Likelihood, Deep Belief Is Not Enough}
	\date{}
	\author{Lucas Theis, Sebastian Gerwinn, Fabian Sinz and Matthias Bethge \\
			Werner Reichardt Centre for Integrative Neuroscience \\
			Bernstein Center for Computational Neuroscience \\
			Max-Planck-Institute for Biological Cybernetics\\
			Spemannstraße 41, 72076 Tübingen, Germany \\
			\texttt{\{lucas,sgerwinn,fabee,mbethge\}@tuebingen.mpg.de}}

	\maketitle

	\begin{abstract}%
		Statistical models of natural stimuli provide an important tool for
		researchers in the fields of machine learning and computational
		neuroscience. A canonical way to quantitatively assess and compare the
		performance of statistical models is given by the likelihood. One class of
		statistical models which has recently gained increasing popularity and
		has been applied to a variety of complex data are deep belief
		networks. Analyses of these models, however, have been typically limited
		to qualitative analyses based on samples due to the computationally
		intractable nature of the model likelihood.  Motivated by these
		circumstances, the present article  provides a consistent estimator for
		the likelihood that is both computationally tractable and simple to
		apply in practice. Using this estimator, a deep belief network which has
		been suggested for the modeling of~natural image patches is
		quantitatively investigated and compared to other models~of natural
		image patches.  Contrary to earlier claims based on qualitative results,
		the results presented in this article provide evidence that the model
		under investigation is not a particularly good model for natural images.
	\end{abstract}

	\vspace{8pt}

	\section{Introduction}
		Statistical models of naturally occurring stimuli constitute an important
		tool in machine learning and computational neuroscience, among many other
		areas. In machine learning, they have been applied both to supervised and
		unsupervised problems, such as denoising \citep[e.g.,][]{Lyu:2007p7424},
		classification \citep[e.g.,][]{Lee:2009p6429} or prediction
		\citep[e.g.,][]{dorettoCWS03ijcv}.  In computational neuroscience, statistical
		models have been used to analyze the structure of natural images as part
		of the quest to understand the tasks faced by the visual system of the
		brain \citep[e.g.,][]{Lewicki:2005p7425,Olshausen:1996}. Other examples
		include generative statistical models studied to derive better models of
		perceptual learning. Underlying these approaches is the assumption that
		the low-level areas of the brain adapt to the statistical structure of
		their sensory inputs and are less concerned with goals of behavior.

		An important measure to assess the performance of a statistical model is
		the likelihood which allows us to objectively compare the density
		estimation performance of different models.  Given two model instances
		with the same prior probability and a test set of data samples, the ratio
		of their likelihoods already tells us everything we need to know to decide
		which of the two models is more likely to have generated the dataset.
		Furthermore, for densities $p$ and $\tilde p$, the negative expected
		log-likelihood represents the \textit{cross-entropy} (or \textit{expected
		log-loss}) term of the Kullback-Leibler~(KL)~divergence,
		\begin{align*}
			D_\text{KL}[\tilde p || p] =  -\sum_x \tilde p(x) \log p(x) - \mathcal{H}[\tilde p],
		\end{align*}
		which is always non-negative and zero if and only if $p$ and $\tilde p$
		are identical. The main motivation for the KL-divergence stems from coding
		theory, where the cross-entropy represents the coding cost of encoding
		samples drawn from~$\tilde p$ with a code that would be optimal for
		samples drawn from~$p$. Correspondingly, the KL-divergence represents the
		additional coding cost created by using an optimal code which assumes the
		distribution of the samples to be $p$ instead of $\tilde p$. Finally, the
		likelihood allows us to directly examine the success of maximum likelihood
		learning for different training settings. Unfortunately, for many
		interesting models, the likelihood is intractable to compute exactly.

\begin{figure}[t]
	\centering
	\includegraphics[width=0.47\textwidth]{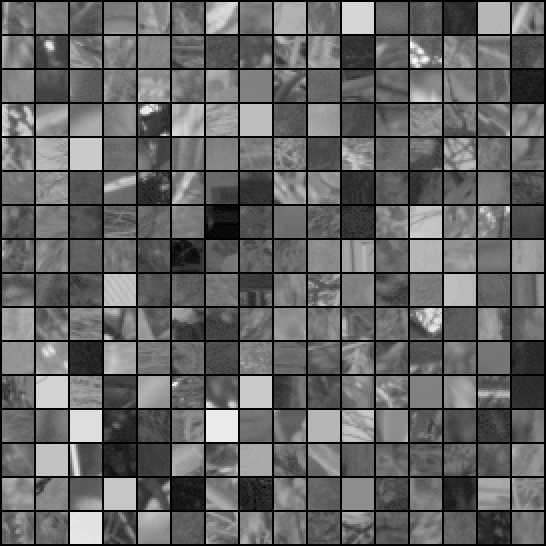}
	\includegraphics[width=0.47\textwidth]{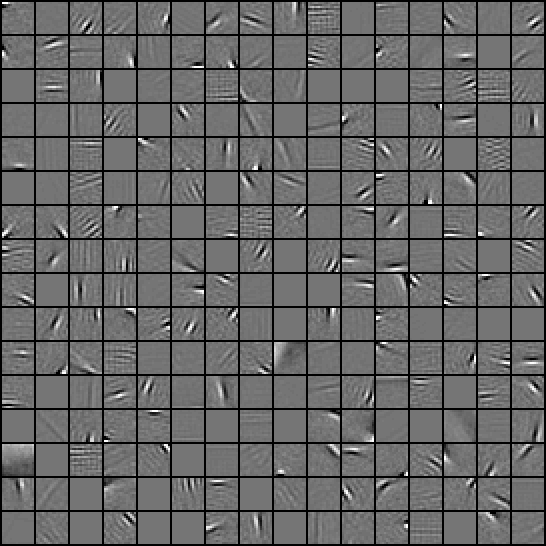}
	\caption{\textit{Left:} Natural image patches sampled from the van Hateren
		dataset \citep{vanHateren:1998p7367}. \textit{Right:} Filters learned by a deep belief network trained on whitened image patches.}
	\label{fignatim}
\end{figure}

		One such class of models which has attracted a lot of attention in recent
		years is given by deep belief networks.  Deep belief networks are
		hierarchical generative models introduced by Hinton et al.
		\citep{Hinton:2006p2883} together with a greedy learning rule as an
		approach to the long-standing challenge of training \textit{deep} neural
		networks, that is, hierarchical neural networks with many layers. In
		supervised tasks, they have been shown to learn representations which can
		be successfully employed in classification tasks, such as character
		recognition~\citep{Hinton:2006p2884} and speech
		recognition~\citep{Mohamed:2009p7395}. In unsupervised tasks, where the
		likelihood is particularly important, they have been applied to a wide
		variety of complex datasets, such as patches of natural images
		\citep{Osindero:2008p6424,Ranzato:2010p7294,Ranzato:2010p7293,Lee:2007p7146},
		motion capture recordings \citep{Taylor:2007p4750} and images of faces
		\citep{Susskind:2008p7394}.

		When applied to natural images, deep belief networks have been shown to
		develop biologically plausible features \citep{Lee:2007p7146} and samples
		from the model were shown to adhere to certain statistical regularities
		also found in natural images \citep{Osindero:2008p6424}. Examples of
		natural image patches and features learned by a deep belief network are
		presented in Figure \ref{fignatim}.

		In this article, after reviewing the relevant aspects of deep belief
		networks, we will derive a consistent estimator for its likelihood and
		demonstrate the estimator's applicability in practice by evaluating a
		model trained on natural image patches. After a thorough quantitative
		analysis, we will argue that the deep belief network under consideration
		is not a particularly good model for estimating the density of small
		natural image patches, as it is outperformed with respect to the
		likelihood even by simple mixture models. Furthermore, we will show that
		adding layers to the network has only a small effect on the overall
		performance of the model if each layer is trained well enough and will
		offer a possible explanation for this observation by analyzing a best-case
		scenario of the greedy learning procedure commonly used for training deep
		belief networks.

	\section{Models}
		In this chapter we will review the statistical models used in the
		remainder of this article.  In particular, we will describe the restricted
		Boltzmann machine (RBM) and some of its variants which constitute the
		main building blocks for constructing deep belief networks (DBNs).
		Furthermore, we will discuss some of the models' properties relevant for
		estimating the likelihood of DBNs. Readers familiar with DBNs might want
		to skip this section or skim it to get acquainted with the notation.

		Throughout this article, the goal of applying statistical models is
		assumed to be the approximation of a particular distribution of
		interest---often called the \textit{data distribution}.  We will denote
		this distribution by $\tilde p$.

		\subsection{Boltzmann Machines}
			\label{secbm}
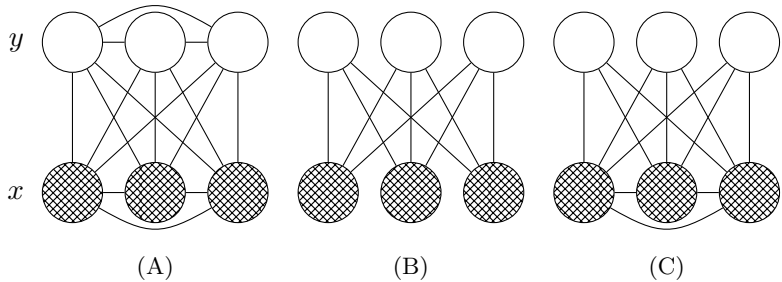
\begin{figure}[h]
	\centering
		\begin{tikzpicture}
			\foreach \x in {0,1,2} {
				\node[hunit] (h\x) at (\x * 1.1, 2) {};
				\node[vunit] (v\x) at (\x * 1.1, 0) {};
			}

			\foreach \x in {0,1,2}
				\foreach \y in {0,1,2}
					\draw [-] (v\x) to (h\y);
				\draw [-] (v0) to (v1);
				\draw [-] (v1) to (v2);
				\draw [-] (h0) to (h1);
				\draw [-] (h1) to (h2);
				\draw [-,bend right,looseness=1.3] (v0) to (v2);
				\draw [-,bend left,looseness=1.3] (h0) to (h2);

			\node [left=0.1cm of v0] {$x$};
			\node [left=0.1cm of h0] {$y$};
			\node [below=0.3cm of v1] {\footnotesize (A)};

			\foreach \x in {0,1,2} {
				\node[hunit] (h\x) at (3.4 + \x * 1.1, 2) {};
				\node[vunit] (v\x) at (3.4 + \x * 1.1, 0) {};
			}

			\foreach \x in {0,1,2}
				\foreach \y in {0,1,2}
					\draw [-] (v\x) to (h\y);

			\node [below=0.3cm of v1] {\footnotesize (B)};

			\foreach \x in {0,1,2} {
				\node[hunit] (h\x) at (6.8 + \x * 1.1, 2) {};
				\node[vunit] (v\x) at (6.8 + \x * 1.1, 0) {};
			}

			\foreach \x in {0,1,2}
				\foreach \y in {0,1,2}
					\draw [-] (v\x) to (h\y);

			\draw [-] (v0) to (v1);
			\draw [-] (v1) to (v2);
			\draw [-,bend right,looseness=1.3] (v0) to (v2);

			\node [below=0.3cm of v1] {\footnotesize (C)};
		\end{tikzpicture}
	\caption{Boltzmann machines with different constraints on their connectivity. Filled nodes denote
	observed variables, unfilled nodes denote hidden variables. \textit{A:} A fully connected latent-variable
	Boltzmann machine. \textit{B:} A restricted Boltzmann machine forming a bipartite graph. \textit{C:} A
	semi-restricted Boltzmann machine, which in contrast to RBMs also allows connections between the visible units.}
	\label{figzoo}
\end{figure}
			A Boltzmann machine is a potentially fully connected \textit{undirected
			graphical model}---or \textit{Markov random field}---with binary random
			variables. Its probability mass function is a Boltzmann distribution
			over $2^k$ binary states $s \in \{0, 1\}^k$ which is defined in terms
			of an \textit{energy function}~$E$,
			\begin{align}
				\label{eq18}
				q(s) = \frac{1}{Z} \exp(-E(s)), \quad Z = \sum_s \exp(-E(s)),
			\end{align}
			where $E$ is given by
			\begin{align}
				\label{eq29}
				E(s) = - \frac{1}{2} s^\top W s - b^\top s = -\frac{1}{2} \sum_{i, j} s_i w_{ij} s_j - \sum_i s_i b_i
			\end{align}
			and depends on a symmetric weight matrix $W \in \mathbb{R}^{k \times
			k}$ with zeros on the diagonal, i.e. $w_{ii} = 0$ for all $i = 1, ...,
			k$, and bias terms $b \in \mathbb{R}^k$. $Z$ is called
			\textit{partition function} and ensures the normalization of $q$. In
			the following, unnormalized distributions will be marked with an
			asterisk:
			\begin{align*}
				q^*(s) = Z q(s) = \exp(-E(s)).
			\end{align*}
			Samples from the Boltzmann distribution can be obtained via
			\textit{Gibbs sampling}, which operates by conditionally sampling each
			univariate random variable until some convergence criterion is reached.
			From definitions \eqref{eq18} and \eqref{eq29} it follows that the
			conditional probability of a unit $i$ being on given the states of all
			other units $s_{j \neq i}$ is given by
			\begin{align*}
				q(s_i = 1 \mid s_{j \neq i}) = g\left( \sum_j w_{ij} s_j + b_i \right),
			\end{align*}
			where $g(x) = 1 / (1 + \exp(-x))$ is the sigmoidal logistic function.
			The Boltzmann machine can be seen as a stochastic generalization of the
			binary Hopfield network, which is based on the same
			energy function but updates its units deterministically using a step
			function, i.e. a unit is set to~1 if $\sum_j w_{ij} s_i + b_i > 0$ and
			set to~0 otherwise.  In the limit of increasingly large weight
			magnitudes, the logistic function becomes a step function and the
			deterministic behavior of the Hopfield network can be recovered with
			the Boltzmann machine \citep{Hinton:2007}.

			Of particular interest for building DBNs are \textit{latent variable
			Boltzmann machines}, that is, Boltzmann machines for which the states
			$s$ are only partially observed (\figref{figzoo}). We will refer to
			states of observed---or visible---random variables as $x$ and to states
			of unobserved---or hidden---random variables as $y$, such that $s$ can
			be written as $s = (x, y)$.

			Approximation of the data distribution $\tilde p(x)$ with the model
			distribution $q(x)$ via maximum likelihood (ML) learning can be
			implemented by following the gradient of the model log-likelihood. In
			Boltzmann machines, this gradient is conceptually simple yet
			computationally very hard to evaluate. The gradient of the expected
			log-likelihood with respect to some parameter $\theta$ of the energy
			function \hbox{is \citep{Salakhutdinov:2009p7274}:}
			\begin{align}
				\label{eq15}
				\mathcal{E}_{\tilde p(x)}\left[\frac{\partial}{\partial \theta} \log q(x)\right]
				&= \mathcal{E}_{q(x, y)}\left[ \frac{\partial}{\partial \theta} E(x, y) \right] - \mathcal{E}_{\tilde p(x) q(y \mid x)}\left[ \frac{\partial}{\partial \theta} E(x, y) \right].
			\end{align}
			The first term on the right-hand side of this equation is the expected
			gradient of the energy function when both hidden and observed states
			are sampled from the model, while the second term is the expected
			gradient of the energy function when the hidden states are drawn from
			the conditional distribution of the model, given a visible state drawn
			from the data distribution.  Following this gradient increases the
			energy of the states which are more likely with respect to the model
			distribution and decreases the energy of the states which are more
			likely with respect to the data distribution. Remember that by the
			definition of the model \eqref{eq18}, states with higher energy are
			less likely than states with lower energy.

			As an example, the gradient of the log-likelihood with respect to the
			weight connecting a visible unit $x_i$ and a hidden unit $y_j$ becomes
			\begin{align*}
				\mathcal{E}_{\tilde p(x) q(y \mid x)}[ x_i y_j ] - \mathcal{E}_{q(x, y)}[ x_i y_j ].
			\end{align*}
			A step in the direction of this gradient can be interpreted as a
			combination of Hebbian and anti-Hebbian learning \citep{Hinton:2003p1},
			where the first term corresponds to Hebbian learning and the second
			term to anti-Hebbian learning, respectively.

			Evaluating the expectations, however, is computationally intractable
			for all but the simplest networks.  Even approximating the expectations
			with Monte Carlo methods is typically very slow \citep{Long:2010p7326}.
			Two measures can be taken to make learning in Boltzmann machines
			feasible: constraining the Boltzmann machine in some way, or replacing
			the likelihood with a simpler objective function. The former approach
			led to the introduction of RBMs, which will be discussed in the next
			section. The latter approach led to the now widely used contrastive
			divergence (CD) learning rule \citep{Hinton:2002p6067} which represents
			a tractable approximation to ML learning: In CD learning, the
			expectation over the model distribution $q(x, y)$ is replaced by an
			expectation over
			\begin{align*}
				q_\text{CD}(x, y) = \sum_{x_0, y_1} \tilde p(x_0) q(y_1 \mid x_0) q(x \mid y_1) q(y \mid x),
			\end{align*}
			from which samples are obtained by taking a sample $x_0$ from the data
			distribution, updating the hidden units, updating the visible units,
			and finally updating the hidden units again, while in each step keeping
			the respective set of other variables fixed.  This corresponds to a
			single sweep of Gibbs sampling through all random variables of the
			model plus an additional update of the hidden units. If instead $n$
			sweeps of Gibbs sampling are used, the learning procedure is generally
			referred to as CD($n$) learning. For $n \rightarrow \infty$, ML
			learning is \hbox{regained \citep{Salakhutdinov:2009p7274}}.

		\subsection{Restricted Boltzmann Machines}
			A restricted Boltzmann machine (RBM) \citep{Smolensky:1986p7136} is a
			Boltzmann machine whose energy function is constrained such that no
			direct interaction between two visible units or two hidden units is
			possible,
			\begin{align*}
				E(x, y) = - x^\top W y - b^\top x - c^\top y.
			\end{align*}
			The corresponding graph has no connections between the visible units
			and no connections between the hidden units and is hence
			bipartite~(\figref{figzoo}). The weight matrix $W \in \mathbb{R}^{m
			\times n}$ is different to the one in equation~\eqref{eq29} in that it
			now only contains interaction terms between the $m$ visible units and
			the $n$ hidden units and therefore no longer needs to be symmetric or
			constrained in any other way. Despite these constraints it has been
			shown that RBMs are universal approximators, i.e. for any distribution
			over binary states and any $\varepsilon > 0$, there exists an RBM with
			a KL-divergence which is smaller than $\varepsilon$
			\citep{LeRoux:2008p7356}.

			In an RBM, the visible units are conditionally independent given the
			states of the hidden units and vice versa. The conditional distribution
			of the hidden units, for instance, is given by
			\begin{align*}
				q(y \mid x) = \prod_j q(y_j \mid x), \quad q(y_j \mid x) = g(w_j^\top x + c_j),
			\end{align*}
			where $g$ is the logistic sigmoidal function and $w_j$ is the $j$-th
			column of $W$. This allows for efficient Gibbs sampling of the model
			distribution (since one set of variables can be updated in parallel
			given the other) and thus for faster approximation of the
			log-likelihood gradient.  Moreover, the unnormalized marginal
			distributions $q^*(x)$ and $q^*(y)$ of RBMs can be computed
			analytically by integrating out the respective other variable. For
			instance, the unnormalized marginal distribution of the visible units
			becomes
			\begin{align}
				\label{eq6}
				q^*(x) = \exp(b^\top x) \prod_j (1 + \exp(w_j^\top x + c_j)).
			\end{align}
			Two other models which can be used for constructing DBNs are the
			\textit{Gaussian RBM} (GRBM) \citep{Salakhutdinov:2009p7274} and the
			\textit{semi-restricted Boltzmann machine} (SRBM)
			\citep{Osindero:2008p6424}. The GRBM employs continuous instead of
			binary visible units (while keeping the hidden units binary) and can
			thus be used to model continuous data. Its energy function is given by
			\begin{align}
				\label{eq42}
				E(x, y) = \frac{1}{2 \sigma^2} ||x - b||^2 - \frac{1}{\sigma} x^\top W y - c^\top y.
			\end{align}
			A somewhat more general definition allows a different $\sigma$ for each
			individual visible unit \citep{Salakhutdinov:2009p7274}.  As for the
			binary Boltzmann machine, training of the GRBM proceeds by following
			the gradient given in equation \eqref{eq15}, or an approximation
			thereof. Its properties are similar to that of an RBM, except that its
			conditional distribution $q(x \mid y)$ is a multivariate Gaussian
			distribution whose mean is determined by the hidden units,
			\begin{align*}
				q(x \mid y) = \mathcal{N}(x; W y + b, \sigma I).
			\end{align*}
			Each state of the hidden units encodes one mean. $\sigma$ controls the
			variance of each Gaussian and is the same for all states of the hidden
			units.  The GRBM can therefore be interpreted as a mixture of an
			exponential number of Gaussian distributions with fixed, isotropic
			covariance and parameter sharing constraints.

			In an SRBM, only the hidden units are constrained to have no direct
			connections to each other while the visible units are unconstrained
			(\figref{figzoo}). Importantly, analytic expressions are therefore only
			available for $q^*(x)$ but not for $q^*(y)$. Furthermore, $q(x \mid y)$
			is no longer factorial. For efficiently training DBNs, conditional
			independence of the hidden units is more important than conditional
			independence of the visible units \citep{Osindero:2008p6424}. This is
			in part due to the fact that the second-term on the right-hand side of
			equation \eqref{eq15} can still efficiently be evaluated if the hidden
			units are conditionally independent, and in part due to the way
			inference is done in DBNs.

		\subsection{Deep Belief Networks}
			\label{secdbn}
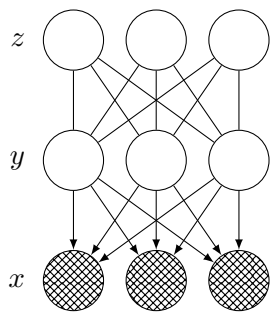
\begin{figure}[h]
	\centering
	\hspace{-0.6cm}
	\begin{tikzpicture}
		\foreach \x in {0,1,2} {
			\node[hunit] (z\x) at (\x * 1.1, 3.2) {};
			\node[hunit] (y\x) at (\x * 1.1, 1.6) {};
			\node[vunit] (x\x) at (\x * 1.1, 0) {};
		}

		\foreach \x in {0,1,2}
			\foreach \y in {0,1,2} {
				\draw [-] (z\x) to (y\y);
				\draw [-latex] (y\x) to (x\y);
			}

		\node [left=0.1cm of x0] {$x$};
		\node [left=0.1cm of y0] {$y$};
		\node [left=0.1cm of z0] {$z$};
	\end{tikzpicture}
	\caption{A graphical model representation of a two-layer deep belief network
	composed of two RBMs. Note that the connections of the first layer are
	directed.}
	\captionsetup{labelsep=colon}
	\label{figdbn}
\end{figure}
			DBNs \citep{Hinton:2006p2883} are hierarchical generative models composed of
			several layers of RBMs or one of their generalizations. While DBNs have been
			widely used as
			part of a heuristic for learning
			multiple layers of feature representations and for pretraining
			multi-layer perceptrons (by initializing the multi-layer perceptron
			with the parameters learned by a DBN), the existence of an efficient
			learning rule has made them become attractive also for density
			estimation tasks.

			For simplicity, we will begin by defining a two-layer DBN. Let $q(x,
			y)$ and $r(y, z)$ be the densities of two RBMs over visible states $x$
			and hidden states $y$ and $z$. Then, the joint probability mass
			function of a two-layer~DBN is defined \hbox{to be}
			\begin{align}
				\label{eq14}
				p(x, y, z) = q(x \mid y) r(y, z).
			\end{align}
			Interestingly, the resulting distribution is best described not as a
			deep Boltzmann machine, as one might expect, or even an undirected
			graphical model, but as a graphical model with undirected connections
			between $y$ and $z$ and directed connections between $x$ and $y$
			(\figref{figdbn}). This characteristic of the model becomes evident in
			the generative process.  A sample from the model can be drawn by first
			Gibbs sampling the distribution $r(y, z)$ of the top layer to produce a
			state for $y$. Afterwards, a sample is drawn from the much simpler
			distribution $q(x \mid y)$.

			The definition can easily be extended to DBNs with three or more layers
			by replacing $r(y, z) = r(y \mid z) r(z)$ with $r(y \mid z) s(z)$,
			where $s(z)$ is the marginal distribution of another RBM. Thus, by
			adding additional layers to the DBN, the prior distribution over the
			top-level hidden units---$r(z)$ for the model defined in equation
			$\eqref{eq14}$---is effectively replaced with a new prior
			distribution---in this case $s(z)$. DBNs with an arbitrary number of
			layers, like RBMs, have been shown to be universal approximators even
			if the number of hidden units in each layer is fixed to the number of
			visible units \citep{Sutskever:2008p6486}.  As mentioned earlier,
			another possibility to generalize DBNs is to allow for more general
			models as layers.  One such model is the SRBM, which can model more
			complex interactions by having less restrictive independce assumptions.
			Alternatively, one could allow for models with units whose conditional
			probability distributions are not just binary, but can be any
			exponential family distribution \citep{Welling:2005p6485}---one
			instance being the GRBM.

			The greedy learning procedure \citep{Hinton:2006p2884} used for
			training DBNs starts by fitting the first-layer RBM (the one closest to
			the observations) to the data distribution.  Afterwards, the prior
			distribution over the hidden units defined by the first layer, $q(y) =
			\sum_x q(x, y)$, is replaced by the marginal distribution of the second
			layer, $r(y)$, and the parameters of the second layer are trained by
			optimizing a lower bound to the log-likelihood of the two-layer DBN. In
			the following, we will derive this lower bound.

			Let $\theta$ be a parameter of $r$. The gradient of the log-likelihood with respect to $\theta$ is
			\begin{align}
				\frac{\partial}{\partial \theta} \log  p(x)
				&= \frac{1}{ p(x)} \sum_{y} \frac{\partial}{\partial \theta}  p(x, y) \nonumber \\
				&= \frac{1}{ p(x)} \sum_{y}  p(x, y) \frac{\partial}{\partial \theta} \log  p(x, y) \nonumber \\
				&= \sum_y  p(y \mid x) \frac{\partial}{\partial \theta} \log(r(y) q(x \mid y)) \nonumber \\
				&= \sum_y  p(y \mid x) \left( \frac{\partial}{\partial \theta} \log r(y) + \underbrace{\frac{\partial}{\partial \theta} \log q(x \mid y)}_{=\ 0} \right) \nonumber \\
				&= \sum_y  p(y \mid x) \frac{\partial}{\partial \theta} \log r(y) \label{eq5}.
			\end{align}
			Approximate ML learning could therefore in principle be implemented by
			training the second layer to approximate the posterior distribution
			\hbox{$p(y \mid x)$} using CD learning or a similar algorithm.
			However, exact sampling from the posterior distribution~\hbox{$p(y \mid
			x)$} is difficult, as its evaluation involves integrating over an
			exponential number of states,
			\begin{align*}
				p(y \mid x) = \frac{p(x, y)}{p(x)} = \frac{q(x \mid y) r(y)}{\sum_y q(x \mid y) r(y)}.
			\end{align*}
			In order to make the training feasbile again, the posterior distribution
			is replaced by the factorial distribution $q(y \mid x)$. Training the DBN in this manner
			optimizes a variational lower bound on the log-likelihood,
			\begin{align}
				\log p(x)
				&= \log \sum_y q(x \mid y) r(y) \nonumber \\
				&= \log \sum_y q(y \mid x) \frac{q(x)}{q(y)} r(y) \label{eq3} \\
				&\geq \sum_y q(y \mid x) \log r(y) + const, \label{eq4}
			\end{align}
			where \eqref{eq3} follows from Bayes' theorem, \eqref{eq4} is due to
			Jensen's inequality and $const$ is constant in $\theta$, as only $r$
			depends on $\theta$. Taking the derivative of \eqref{eq4} with respect
			to $\theta$ yields $\eqref{eq5}$ with the posterior distribution $p(y
			\mid x)$ replaced by $q(y \mid x)$. The greedy learning procedure can
			be generalized to more layers by training each additional layer to
			approximate the distribution obtained by conditionally sampling from
			each layer in turn, starting with the lowest layer.

	\section{Likelihood Estimation}
		In this section, we will discuss the problem of estimating the likelihood of a two-layer DBN
		with joint density
		\begin{align}
			\label{eq20}
			p(x, y, z) = q(x \mid y) r(y, z).
		\end{align}
		That is, for a given visible state $x$, to estimate the value of
		\begin{align*}
			p(x) = \sum_{y, z} q(x \mid y) r(y, z).
		\end{align*}
		As we will see later, this problem can easily be generalized to more
		\hbox{layers. As} before, $q(x, y)$ and $r(y, z)$ refer to the densities
		of two RBMs.

		Two difficulties arise when dealing with this problem in the context of
		DBNs. First, $r(y, z)$ depends on a partition function $Z_r$ whose exact
		evaluation requires integration over an exponential number of states.
		Second, despite our ability to integrate analytically over $z$, even
		computing just the unnormalized likelihood still requires integration over
		an exponential number of hidden states $y$,
		\begin{align*}
			p^*(x) = \sum_y q(x \mid y) r^*(y).
		\end{align*}
		After briefly reviewing previous approaches to resolving these
		difficulties, we will propose an unbiased estimator for $p^*(x)$, its
		contribution being a possible solution to the second problem, and discuss
		how to construct a consistent estimator for $p(x)$ based on this
		estimator. Finally, we will demonstrate its applicability to more general
		DBNs.

		\subsection{Previous Work}
			\subsubsection{Annealed Importance Sampling}
				\citet{Salakhutdinov:2008p7269} have shown how \textit{annealed
				importance sampling} (AIS) \citep{Neal:2001p7298} can be used to
				estimate the partition function of a restricted Boltzmann machine.
				Since our estimator will also rely on AIS estimates of the partition
				function, we will shortly describe the procedure here.

				Importance sampling is a Monte Carlo method for unbiased estimation
				of expectations \citep{JCMacKay:2003p7080} and is based on the
				following observation: Let $s$ be a density with $s(x) > 0$ whenever
				$q^*(x) > 0$ and let $w(x) = \frac{q^*(x)}{s(x)}$, then
				\begin{align}
					\label{eq11}
					\sum_x q^*(x) f(x) = \sum_x s(x) \frac{q^*(x)}{s(x)} f(x) = \mathcal{E}_{s(x)}\left[w(x) f(x)\right]
				\end{align}
				for any function $f(x)$. $s$ is called the \textit{proposal
				distribution} and $w(x)$ is called \textit{importance weight}. For
				$f(x) = 1$, we get
				\begin{align}
					\label{eq10}
					\mathcal{E}_{s(x)}\left[w(x)\right] = \sum_x s(x) \frac{q^*(x)}{s(x)} = Z_q.
				\end{align}
				Estimates of the partition function $Z_q$ can therefore be obtained
				by drawing samples $x^{(n)}$ from a proposal distribution and
				averaging the resulting importance weights $w(x^{(n)})$. It was
				pointed out in \citep{Minka:2005p7350} that minimizing the variance
				of the importance sampling estimate of the partition function
				\eqref{eq10} is equivalent to minimizing an
				$\alpha$-divergence\footnote{With $\alpha = 2$. $\alpha$-divergences
				are a generalization of the KL-divergence.} between the proposal
				distribution $s$ and the true distribution $q$. Therefore, for the
				estimate to work well in practice, $s$ should be both close to $q$
				and easy to sample from.

				Annealed importance sampling \citep{Neal:2001p7298} tries to
				circumvent some of the problems associated with finding a suitable
				proposal distribution.  Assume we can construct a distribution $s_1$
				which approximates $q$ well, but which is still difficult to sample
				from or which we can only evaluate up to a normalization factor. Let
				$s_2$ be another distribution. This distribution will effectively
				act as a proposal distribution for $s_1$. Further, let $T_1$ be a
				\textit{transition operator} which leaves the distribution of $s_1$
				invariant, i.e. let $T_1(x_0; x_1)$ be a probability distribution
				over $x_0$ depending on $x_1$, such that
				\begin{align*}
					s_1(x_0) = \sum_{x_1} s_1(x_1) T_1(x_0; x_1).
				\end{align*}
				We then have
				\begin{align*}
					Z_q 
					&= \sum_{x_0} s_1(x_0) \frac{q^*(x_0)}{s_1(x_0)} \\
					&= \sum_{x_0} \sum_{x_1} s_1(x_1) T_1(x_0; x_1) \frac{q^*(x_0)}{s_1(x_0)} \\
					&= \sum_{x_0} \sum_{x_1} s_2(x_1) T_1(x_0; x_1) \frac{s_1^*(x_1)}{s_2(x_1)} \frac{q^*(x_0)}{s_1^*(x_0)}.
				\end{align*}
				Note that we don't have to know the partition function of $s_1$ to
				evaluate the right-hand term. Also note that we don't need to sample
				from $s_1$ but only from $T_1$ if we want to estimate this term via
				Monte Carlo integration.  If $s_2$ is still too difficult to handle,
				we can apply the same trick again by introducing a third
				distribution $s_3$ and a transition operator $T_2$ for $s_2$. By
				induction, we can see that
				\begin{align*}
					Z_q
					&= \sum_x s_n(x_{n - 1}) T_{n - 1}(x_{n - 2}; x_{n - 1}) \cdots T_1(x_0; x_1) \frac{s_{n - 1}^*(x_{n - 1})}{s_n(x_{n - 1})} \cdots \frac{q^*(x_0)}{s_1^*(x_0)}, 
				\end{align*}
				where the sum integrates over all $x = (x_0, ..., x_{n - 1})$.
				Hence, in order to estimate the partition function, we can draw
				independent samples $x_{n - 1}$ from a simple distribution $s_n$,
				use the transition operators to generate intermediate samples
				$x_{n - 2}, ..., x_0$, and use the product of fractions in the
				preceding equation to compute importance weights, which we then
				average.

				In order to be able to apply AIS to RBMs, a sequence of intermediate
				distributions and corresponding \textit{annealing weights} is
				defined:
				\begin{align*}
					s_k^*(x) = q^*(x)^{1 - \beta_k} s(x)^{\beta_k}, \quad \beta_k \in [0, 1]
				\end{align*}
				for $k = 0, ..., n$, where $\beta_0 = 0$ and $\beta_n = 1$. If we
				also choose an RBM for $s$, then $s_k$ is itself a Boltzmann
				distribution whose energy function is a weighted sum of the energy
				functions of $s$ and $q$. Similarly, natural and efficient
				implementations based on Gibbs sampling can be found for the
				transition \hbox{operators $T_k$}.

			\subsubsection{Estimating Lower Bounds}
				In \citep{Salakhutdinov:2008p7269} it was also shown how estimates of a
				lower bound on the log-likelihood,
				\begin{align}
					\label{eq19}
					\log p(x) 
					&\geq \sum_y q(y \mid x) \log \frac{r^*(y) q(x \mid y)}{q(y \mid x)} - \log Z_r \\
					\label{eq21}
					&= \sum_y q(y \mid x) \log r^*(y) q(x \mid y) + \mathcal{H}[q(y \mid x)] - \log Z_r,
				\end{align}
				can be obtained, provided the partition function $Z_r$ is given.
				This is the same lower bound as the one optimized during greedy learning \eqref{eq4}.
				Since $q(y \mid x)$ is factorial, the entropy $\mathcal{H}[q(y \mid x)]$ can be
				computed analytically. The only term which still needs to be
				estimated is the first term on the right-hand side of equation
				\eqref{eq21}. This was achieved in \citep{Salakhutdinov:2008p7269}
				by drawing samples from~$q(y \mid x)$.

			\subsubsection{Consistent Estimates}
				In \citep{Murray:2009p7319}, carefully designed Markov chains were
				constructed to give unbiased estimates for the inverse posterior
				probability $\frac{1}{p(y \mid x)}$ of some fixed hidden state $y$.
				These estimates were then used to get unbiased estimates of $p^*(x)$
				by taking advantage of the fact $p^*(x) = \frac{p^*(x, y)}{p(y \mid
				x)}$.  The corresponding partition function was estimated using AIS,
				leading to an overall estimate of the likelihood that tends to
				overestimate the true likelihood. While the estimator was
				constructed in such a way that even very short runs of the Markov
				chain result in unbiased estimates of~$p^*(x)$, even a single step
				of the Markov chain is slow compared to sampling from $q(y \mid x)$
				as it was done for the estimation of the lower bound~\eqref{eq19}.

		\subsection[A New Estimator for Deep Belief Networks]{A New Estimator for DBNs}
			The estimator we will introduce in this section shares the same formal
			properties as the estimator proposed in \citep{Murray:2009p7319}, but
			will utilize samples drawn from $q(y \mid x)$. This will make it
			conceptually as simple and as easy to apply in practice as the
			estimator for the lower bound \eqref{eq19}, while providing us with
			consistent estimates of~$p(x)$.

			\subsubsection[Definition of the Estimator]{Definition}
				Let $p(x, y, z)$ be the joint density of a DBN as defined in
				equation \eqref{eq20}. By applying Bayes' theorem, we obtain
				\begin{align}
					p(x)
					&= \sum_y q(x \mid y) r(y) \\
					&= \sum_y q(y \mid x) \frac{q(x)}{q(y)} r(y) \\
					\label{eq1}
					&= \sum_y q(y \mid x) \frac{q^*(x)}{q^*(y)} \frac{r^*(y)}{Z_r}.
				\end{align}
				An obvious choice for an estimator of $p(x)$ is then
				\begin{align}
					\label{eq2}
					\hat p_N(x)
					&= \frac{1}{N} \sum_n \frac{q^*(x)}{q^*(y^{(n)})} \frac{r^*(y^{(n)})}{Z_r}  \\
					&= q^*(x) \frac{1}{Z_r N} \sum_n \frac{r^*(y^{(n)})}{q^*(y^{(n)})}
				\end{align}
				where $y^{(n)} \sim q(y^{(n)} \mid x) \text{ for } n = 1, ..., N$.
				For RBMs, the unnormalized marginals $q^*(x), q^*(y)$ and $r^*(y)$
				can be computed analytically \eqref{eq6}. Note that the partition
				function $Z_r$ only has to be calculated once for all visible states
				we wish to evaluate. Intuitively, the estimation process can be
				imagined as first assigning a basic value to $x$ using the
				distribution of the first layer, and then with every sample
				adjusting this value depending on how the second layer distribution
				relates to the first layer distribution.

			\subsubsection[Properties of the Estimator]{Properties}
				Under the assumption that the partition function $Z_r$ is known,
				$\hat p(x)$ provides an unbiased estimate of $p(x)$ since the
				sample average is always an unbiased estimate of the expectation.
				However, $Z_r$ is generally intractable to compute exactly so that
				approximations become necessary. In fact, in \citep{Long:2010p7326}
				it was shown that already approximating the partition function of an
				RBM to within a multiplicative factor is generally NP-hard in the
				number of parameters of the RBM.

				If in the estimate \eqref{eq2}, $Z_r$ is replaced by an unbiased estimate $\hat Z_r$,
				then the overall estimate will tend to overestimate the true likelihood,
				\begin{align*}
					\mathcal{E}\left[ \frac{\hat p_N^*(x)}{\hat Z_r} \right]
					&= \mathcal{E}\left[ \frac{1}{\hat Z_r}\right] \mathcal{E}\left[\hat p_N^*(x) \right] \\
					&\geq \frac{1}{\mathcal{E}\left[\hat Z_r\right]} p^*(x)
					= p(x),
				\end{align*}
				where $\hat p_N^*(x) = Z_r \hat p_N(x)$ is an unbiased estimate of
				the unnormalized density.  The second step is a consequence of
				Jensen's inequality and the averages are taken with respect to~$\hat
				p_N(x)$ and~$\hat Z_r$, which are independent; $x$ is held fix.

				While the estimator loses its unbiasedness for unbiased estimates of
				the partition function, it still retains its consistency.  Since
				$p_N^*(x)$ is unbiased for all $N \in \mathbb{N}$, it is also
				asymptotically unbiased,
				\begin{align*}
					\plim_{N \rightarrow \infty} p_N^*(x) = p^*(x).
				\end{align*}
				Furthermore, if $\hat Z_{r,N}$ for $N \in \mathbb{N}$ is a consistent sequence
				of estimators for the partition function, it follows that
				\begin{align*}
					\plim_{N \rightarrow \infty} \frac{\hat p^*_N(x)}{\hat Z_{r, N}}
					= \frac{\displaystyle \plim_{N \rightarrow \infty} \hat p^*_N(x)}{\displaystyle \plim_{N \rightarrow \infty} \hat Z_{r, N}}
					= \frac{p^*(x)}{Z_r} = p(x).
				\end{align*}
				Unbiased and consistent estimates of $Z_r$ can be obtained using AIS
				\citep{Salakhutdinov:2008p7269}.  Note that although the estimator
				tends to overestimate the true likelihood in expectation and is
				unbiased in the limit, it is still possible for it to underestimate
				the true likelihood most of the time. This behavior can occur if the
				distribution of estimates is heavily skewed.

				Another question which remains is whether the estimator is good in
				terms of efficiency, or in other words: How many samples are
				required before a reliable estimate of the true likelihood is
				achieved? To address this question, we reformulate the expectation
				in equation \eqref{eq1} to give
				\begin{align*}
					p(x) &= \sum_y q(y \mid x) \frac{p(x, y)}{q(y \mid x)}.
				\end{align*}
				In this formulation it becomes evident that estimating $p(x)$ is
				equivalent to estimating the partition function of $p(y \mid x)$
				using importance sampling. To see this, notice that, for a fixed
				$x$, $p(x, y)$ is just an unnormalized version of $p(y \mid x)$,
				where $p(x)$ is the normalization constant,
				\begin{align*}
					p(y \mid x) = \frac{p(x, y)}{p(x)}.
				\end{align*}
				The proposal distribution in this case is $q(y \mid x)$.  As
				mentioned earlier, the efficiency of importance sampling estimates
				depends on how well the proposal distribution approximates the true
				distribution. Therefore, for the proposed estimator to work well in
				practice, $q(y \mid x)$ should be close to $p(y \mid x)$. Note that
				a similar assumption is made when optimizing the variational lower
				bound \eqref{eq4} during greedy learning.

			\subsubsection{Generalizations}
				The definition of the estimator for two-layer DBNs readily extends
				to DBNs with $L$ layers. If $p(x)$ is the marginal density of a DBN
				whose layers are constituted by RBMs with densities $q_1, ..., q_L$
				and partition functions $Z_1, ..., Z_L$, and if we refer to the
				states of the random vectors in each layer by $x_0, ..., x_L$, where
				$x_0$ contains the visible states and $x_L$ contains the states of
				the top hidden layer, then
				\begin{align*}
					p(x_0) 
					&= \sum_{x_1, ..., x_{L - 1}} q_L(x_{L - 1}) \prod_{l = 1}^{L - 1} q_l(x_{l - 1} \mid x_{l}) \\
					&= \sum_{x_1, ..., x_{L - 1}} q_L(x_{L - 1}) \prod_{l = 1}^{L - 1} q_l(x_{l} \mid x_{l - 1}) \frac{q_l^*(x_{l - 1})}{q_l^*(x_{l})} \\
					&= q_1^*(x_0) \frac{1}{Z_L} \sum_{x_1, ..., x_{L - 1}} \prod_{l = 1}^{L - 1} q_l(x_l \mid x_{l - 1}) \frac{q_{l + 1}^*(x_{l})}{q_{l}^*(x_l)}.
				\end{align*}
				In order to estimate this term, hidden states $x_1, ..., x_L$ are
				generated in a feed-forward manner using the conditional
				distributions $q_l(x_{l}~\mid~x_{l - 1})$.  The weights $\frac{q_{l
				+ 1}^*(x_{l})}{q_l^*(x_{l})}$ are computed along the way, then
				multiplied together and finally averaged over all drawn states.

				Often, a DBN not only contains RBMs but also more general
				distributions $q(x, y)$ \citep[see, for
				example,][]{Roux:2010p7372,Osindero:2008p6424,Ranzato:2010p7294,Ranzato:2010p7293}.
				In this case, analytical expressions of
				the unnormalized distribution over the hidden states $q^*(y)$ might
				be unavailable, as, for example, for the SRBM.  If AIS or some other
				importance sampling method is used for the estimation of the
				partition function, however, the same importance samples and
				importance weights can be used in order to get unbiased estimates of
				$q^*(y)$, as we will show in the following.

				As in equation \eqref{eq11}, let $s$ be a proposal distribution and
				$w$ be importance weights such that
				\begin{align*}
					\sum_x s(x) w(x) f(x) = \sum_x q^*(x) f(x).
				\end{align*}
				for any function $f$. By noticing that $q^*(y) = \sum_x q^*(x) q(y
				\mid x)$, it easy to see how estimates of $q^*(y)$ can be obtained
				using the same importance samples and importance weights which are
				used for estimating the partition function,
				\begin{align*}
					q^*(y) \approx \frac{1}{N} \sum_n w^{(n)} q(y \mid x^{(n)}).
				\end{align*}
				As for the partition function, the importance weights only have to
				be generated once for all $x$ and all hidden states that are part of
				the evaluation. Estimating $q^*(y)$ in this manner, however,
				introduces further bias into the estimator. Also note that a good
				proposal distribution for estimating the partition function need not
				be a good proposal distribution for estimating the marginals. The
				optimal proposal distribution for estimating the marginals would be
				$q(x)$, as in this case any importance weight would take on the
				value of the partition function itself \eqref{eq10}. The optimal
				proposal distribution for estimating the value of the unnormalized
				marginal distribution $q^*(y)$, on the other hand, is $q(x \mid y)$,
				which unfortunately depends on $y$. Therefore, more importance
				samples will be needed in order to get reliable estimates of the
				marginals.

		\subsection{Potential Log-Likelihood}
			\label{secpotential}
			In this section, we will discuss the concept of the \textit{potential
			log-likelihood}---a concept which appears in \citep{LeRoux:2008p7356}.
			By considering a best-case scenario, the potential log-likelihood gives
			an idea of the log-likelihood that can at best be achieved by training
			additional layers using greedy learning. Its usefulness will become
			apparent in the experimental section.

			Let $q(x, y)$ be the distribution of an already trained RBM or one of
			its generalizations, and let $r(y)$ be a second distribution---not
			necessarily the marginal distribution of any Boltzmann machine. As in
			section \ref{secdbn}, $r(y)$ serves to replace the prior distribution
			over the hidden variables, $q(y)$, and thereby improve the marginal
			distribution over \hbox{$x$, $\sum_y q(x \mid y) r(y)$}.  As above, let
			$\tilde p(x)$ denote the data distribution.  Our goal, then, is to
			increase the expected log-likelihood of the model distribution with
			respect to~$r$,
			\begin{align}
				\label{eq12}
				\sum_x \tilde p(x) \log \sum_y q(x \mid y) r(y).
			\end{align}
			In applying the greedy learning procedure, we try to reach this goal by optimizing a lower bound on the
			log-likelihood \eqref{eq4}, or equivalently, by minimizing the following KL-divergence:
			\begin{align*}
				D_\text{KL}\left[ \sum_x \tilde p(x) q(y \mid x) || r(y) \right] &
				= -\sum_x \tilde p(x) \sum_y q(y \mid x) \log r(y) + const,
			\end{align*}
			where \textit{const} is constant in $r$.

			The KL divergence is minimal if $r(y)$ is equal to
			\begin{align}
				\label{eq50}
				\sum_x \tilde p(x) q(y \mid x)
			\end{align}
			for every $y$.
			Since RBMs are universal approximators \citep{LeRoux:2008p7356},
			this distribution could in principle be approximated arbitrarily well by a single,
			potentially very large RBM (provided the~$y$ are binary).

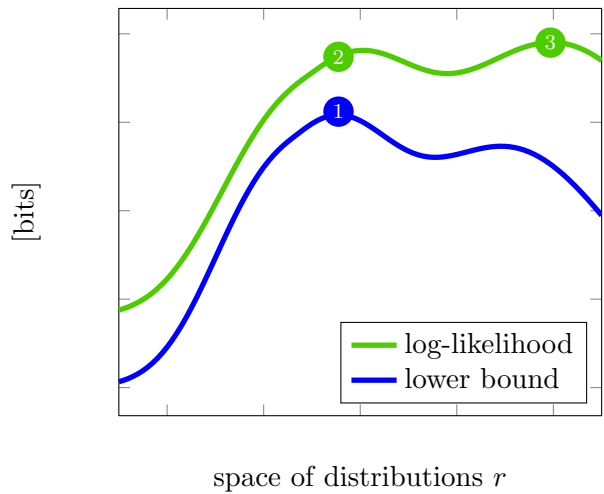
\begin{figure}[t]
	\centering
	\hspace{-1.4cm}
	\begin{tikzpicture}
		\begin{axis}[
				at={(6cm, 0)},
				width=8cm,
				height=7cm,
				enlarge x limits=false,
				xticklabels=none,
				yticklabels=none,
				xlabel={space of distributions $r$},
				ylabel={[bits]},
				legend style={
					at={(0.97, 0.03)},
					anchor=south east,
					cells={anchor=west}
				}
			]

			\addplot[line4, line width=2pt, samples=100] plot {
				0.8 * exp(1 - abs(x + 1)^2.5 / 9) + 1.14 * exp(1 - abs(x - 4)^2 / 15) + 0.8
			};

			\addlegendentry{log-likelihood};
			\addplot[line1, line width=2pt, samples=100] plot {
				0.84 * exp(1 - abs(x + 1.3)^2.5 / 7) + exp(1 - abs(x - 3)^2 / 12)
			};
			\addlegendentry{lower bound};

			\node[font=\scriptsize, circle, draw, line width=2pt, line1, fill=line1, inner sep=1] at (axis cs: -0.45, 3.12) { \color{white} 1 };
			\node[font=\scriptsize, circle, draw, line width=2pt, line4, fill=line4, inner sep=1] at (axis cs: -0.45, 3.74) { \color{white} 2 };
			\node[font=\scriptsize, circle, draw, line width=2pt, line4, fill=line4, inner sep=1] at (axis cs: 3.94, 3.9) { \color{white} 3 };
		\end{axis}
	\end{tikzpicture}
	\caption{A cartoon explaining the potential log-likelihood. The potential log-likelihood is
	the log-likelihood evaluated at~(2), where the lower bound reaches its optimum~(1). This does not
	exclude the existence of a distribution~$r$ for which the log-likelihood is larger than
	the potential log-likelihood, as in (3), but it is unlikely that this point will be found by greedy learning,
	which optimizes~$r$ only with respect to the lower bound.}
	\label{figcartoon}
\end{figure}
			Assume that we have found this distribution, that is, we have maximized
			the lower bound with respect to all possible distributions~$r$. The
			distribution for the DBN which we obtain by replacing $r$ in
			\eqref{eq12} with \eqref{eq50} is then given by
			\begin{align*}
				\sum_y q(x \mid y) \sum_{x_0} \tilde p(x_0) q(y \mid x_0)
				&= \sum_{x_0} \tilde p(x_0) \sum_y q(x \mid y) q(y \mid x_0) \\
				&= \sum_{x_0} \tilde p(x_0) q_0(x \mid x_0),
			\end{align*}
			where we have used the \textit{reconstruction distribution}
			\begin{align*}
				q_0(x \mid x_0) = \sum_y q(x \mid y) q(y \mid x_0),
			\end{align*}
			which can be sampled from by conditionally sampling a state for the
			hidden units, and then, given the state of the hidden units,
			conditionally sampling a \textit{reconstruction} of the visible units.
			The log-likelihood we achieve with this lower-bound optimal
			distribution is given by
			\begin{align}
				\label{eq25}
				\sum_x \tilde p(x) \log \sum_{x_0} \tilde p(x_0) q_0(x \mid x_0).
			\end{align}
			we will refer to this log-likelihood as the \textit{potential
			log-likelihood} (and to the corresponding log-loss as the
			\textit{potential log-loss}). Note that the potential log-likelihood is
			not a true upper bound on the log-likelihood that can be achieved with
			greedy learning, as suboptimal solutions with respect to the lower
			bound might still give rise to higher log-likelihoods.  However, if
			such a solution was found, it would have rather been by accident than
			by design. The situation is depicted in the cartoon in
			\hbox{Figure \ref{figcartoon}}.

	\section{Experiments}
		\label{secexp}
		In order to test the estimator, we considered the task of modeling 4x4
		natural image patches sampled from the van Hateren dataset
		\citep{vanHateren:1998p7367}. We chose a small patch size to allow for
		a more thorough analysis of the estimator's behavior and the effects of
		certain model parameters. In all experiments, a standard battery of
		preprocessing steps was applied to the image patches, including a
		log-transformation, a centering step and a whitening step.
		Additionally, the DC component was projected out and only the other 15
		components of each image patch were used for training
		\cite[for details, see][]{Eichhorn:2009p6426}.

		In \citep{Osindero:2008p6424}, a three-layer DBN based on GRBMs and
		SRBMs was suggested for the modeling of natural image patches. The
		model employed a GRBM in the first layer and SRBMs in the second and
		third layer. In contrast to samples from the same model without lateral
		connections, samples from the proposed model were shown to possess some
		of the statistical regularities also found in natural images, such as
		sparse distributions of pixel intensities and the right pair-wise
		statistics of Gabor filter responses. Furthermore, the first layer of
		the model was shown to develop oriented edge filters
		(\figref{fignatim}). In the following, we will further analyse this
		type of model by estimating its likelihood.

		For training and evaluation, we used 10 independent pairs of training
		and test sets containing 50000 samples each. We trained the models
		using the greedy learning procedure described in Section~\ref{secdbn}.
		The scale-parameter $\sigma$ of the GRBM \eqref{eq42} was chosen via
		cross-validation. After training a GRBM, we initialized the
		second-layer SRBM such that its visible marginal distribution is equal
		to the hidden marginal distribution of the GRBM.  Initializing the
		second layer in this manner has the following advantages. First, after
		initialization, the likelihood of the two-layer DBN consisting of the
		trained GRBM and the initialized SRBM is equal to the likelihood of the
		GRBM.  Second, the lower bound on the DBN's log-likelihood \eqref{eq4}
		is equal to its actual log-likelihood. Using the notation of the
		previous sections:
		\begin{align*}
			r(y) = q(y) \Rightarrow \sum_y q(y \mid x) \log \left(r(y) \frac{q(x)}{q(y)}\right) = \log p(x).
		\end{align*}
		As a consequence, an improvement in the lower bound necessarily leads
		to an improvement in the log-likelihood \citep{Salakhutdinov:2009p7274}.

		All trained models were evaluated using the proposed estimator. We used
		AIS in order to estimate the partition functions and the marginals of
		the SRBMs. Performances were measured as average log-loss in bits and
		normalized by the number of components. Details on the training and
		evaluation parameters can be found in Appendix A.

		\subsection{Small Scale Experiment}
			\begin{table}[t]
				\centering
				\begin{tabular}{|c|c|c|}
					\hline
					layers & true avg. log-loss & est. avg. log-loss \\
					\hline
					1 & 2.0550777 & 2.0551289 \\
					2 & 2.0550775 & 2.0550734 \\
					3 & 2.0550773 & 2.0544256 \\
					\hline
				\end{tabular}
				\caption{True and estimated log-loss of a small-scale version of the model. Adding
				more layers to the network does not help to improve the performance if the GRBM employs only
				few hidden units.}
				\label{tblbrute}
			\end{table}

			In a first experiment, we investigated a small-scale version of the
			model for which the likelihood is still tractable. It employed 15
			hidden units in the first layer, 15 hidden units in the second layer
			and 50 hidden units in the third layer, where each layer was trained
			for 50 epochs using CD(1).  Brute-force and estimated results are
			given in Table \ref{tblbrute}.

			A first observation which can be made is that the estimated
			performance is very close to the true performance. Another
			observation is that the second and third layer do not help to
			improve the performance of the model, which hints at the fact that
			the 15 hidden units of the GRBM are unable to capture much of the
			information in the continuous visible units.

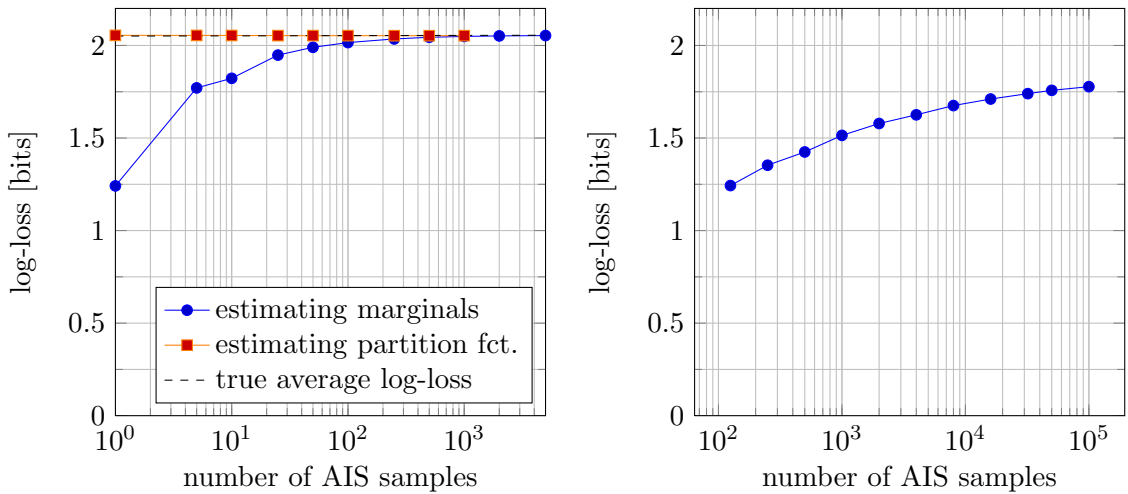
\begin{figure}[t]
	\centering
	\begin{tikzpicture}
		\begin{semilogxaxis}[
				at={(0,0)},
				height=7cm,
				width=7.3cm,
				ymin=0.0,
				ymax=2.2,
				xlabel={number of AIS samples},
				ylabel={log-loss [bits]},
				enlarge x limits=false,
				minor y tick num=1,
				ymajorgrids=true,
				yminorgrids=true,
				xmajorgrids=true,
				xminorgrids=true,
				legend style={
					at={(0.97, 0.03)},
					cells={anchor=west},
					anchor=south east
				}
			]
			\addplot plot[blue] coordinates {
				(1, 1.2413)
				(5, 1.7708)
				(10, 1.8225)
				(25, 1.9480)
				(50, 1.9901)
				(100, 2.0156)
				(250, 2.0352)
				(500, 2.0442)
				(1000, 2.0489)
				(2000, 2.0516)
				(5000, 2.0535)
			};
			\addlegendentry{estimating marginals}

			\addplot plot[orange] coordinates {
				(1, 2.0547)
				(5, 2.0543)
				(10, 2.0540)
				(25, 2.0532)
				(50, 2.0531)
				(100, 2.0531)
				(250, 2.0528)
				(500, 2.0528)
				(1000, 2.0529)
			};
			\addlegendentry{estimating partition fct.}

			\addplot plot[black, dashed, no markers] coordinates {
				(1, 2.05077)
				(5000, 2.055077)
			};
			\addlegendentry{true average log-loss}
		\end{semilogxaxis}

		\begin{semilogxaxis}[
				at={(7.7cm,0)},
				height=7cm,
				width=7.3cm,
				xlabel={number of AIS samples},
				ylabel={log-loss [bits]},
				ymax=2.2,
				ymin=0.0,
				minor y tick num=1,
				ymajorgrids=true,
				yminorgrids=true,
				xmajorgrids=true,
				xminorgrids=true,
				legend style={
					cells={anchor=west}
				}
			]

			\addplot plot [blue] coordinates {
				(125,   1.24309083481)
				(250,   1.35304103235)
				(500,   1.42462139743)
				(1000,  1.51415314231)
				(2000,  1.57841541436)
				(4000,  1.62529246712)
				(8000,  1.67530034976)
				(16000, 1.71064852704)
				(32000, 1.73968235915)
				(50000, 1.75751824531)
				(100000, 1.77739326402)
			};
		\end{semilogxaxis}
	\end{tikzpicture}
	\caption{\textit{Left:} A small-scale DBN was evaluated while either only estimating
			the partition function of the third-layer SRBM (orange curve) or
			estimating the hidden marginal distribution of the second-layer SRBM
			(blue curve), while using different numbers of AIS samples. The
			parameters of the AIS procedure were the same for both estimates. In
			particular, the same number of intermediate annealing distributions
			was used.  Unsurprisingly, the estimated log-loss is more sensitive
			to the number of samples used for estimating the marginals.
			\textit{Right:} The graph shows the estimated performance of DBN-100
			while changing the number of importance samples used to estimate the
			marginals of the second-layer SRBM. The plot indicates that the true
			log-loss is still slightly larger than the estimates we obtained
			even after taking $10^5$ samples.}
	\label{figis}
\end{figure}
			In order to evaluate the likelihood of this model using the proposed
			estimator, the unnormalized marginals of the second-layer SRBM's
			hidden units with respect to the SRBM as well as the partition
			function of the third layer SRBM had to be estimated.  We
			investigated the effect of the number of importance samples used in
			both estimates on the overall estimate of the log-loss and made the
			following observations. First, almost no error could be observed in
			the estimates of the partition function---and hence of the
			log-loss---even if just one importance sample was used (left plot in
			\figref{figis}). This is the case if the proposal distribution is
			very close to the true distribution, as can be seen from equation
			\eqref{eq10} by replacing the former with the latter. However, the
			reason for this observation is likely to be found in the small model
			size and the fact that the third layer contributes virtually nothing
			to an explanation of the data. As the model becomes larger, more
			samples will be required. Second, as expected, many more samples are
			needed for a satisfactory approximation of the marginals. Using too
			few samples led to overestimation of the likelihood and
			underestimation of the log-loss, respectively.

		\subsection{Model Comparison}
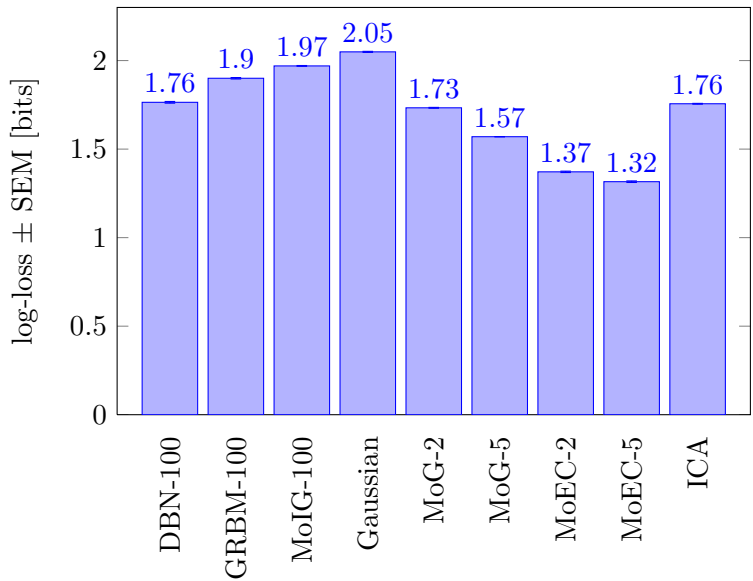
\begin{figure}[t]
	\centering
	\hspace{-1.2cm}
	\begin{tikzpicture}
		\begin{axis}[
				ylabel={log-loss $\pm$ SEM [bits]},
				ybar,
				width=10cm,
				height=7cm,
				bar width=21pt,
				xtick={0,1,2,3,4,5,6,7,8},
				xtick style={color=white},
				xticklabels={
					DBN-100,
					GRBM-100,
					MoIG-100,
					Gaussian,
					MoG-2,
					MoG-5,
					MoEC-2,
					MoEC-5,
					ICA
				},
				ymin=0,
				ymax=2.3,
				x tick label style={rotate=90,anchor=east},
				nodes near coords,
				point meta=y
			]
			\addplot plot[error bars/.cd,y dir=both,y explicit] coordinates {
				(0, 1.76428) +- (0, 4.9E-03)
				(1, 1.9001) +- (0, 4.1E-03)
				(2, 1.9698) +- (0, 2.4E-03)
				(3, 2.0492) +- (0, 3.0E-03)
				(4, 1.7332) +- (0, 2.5E-03)
				(5, 1.5700) +- (0, 7.4E-04)
				(6, 1.3716) +- (0, 4.0E-03)
				(7, 1.3160) +- (0, 4.5E-03)
				(8, 1.7561) +- (0, 2.3E-03)
			};
		\end{axis}
	\end{tikzpicture}
	\caption{A comparison of different models. For each model, the estimated log-loss in
	bits per data component is shown, averaged over 10 independent trials with independent training and test
	sets. The number behind each model hints either at the number of hidden units or at the
	number of mixture components used. All GRBMs and DBNs were trained with CD(1). Larger values
	correspond to worse performance.}
	\label{figcomparison}
\end{figure}
			In a next experiment, we compared the performance of a larger
			instantiation of the model to the performance of linear ICA
			\citep{Eichhorn:2009p6426} as well as several mixture distributions.
			The model employed 100 hidden units in each layer and each layer was
			trained for 100 epochs. As in \citep{Osindero:2008p6424}, CD(1) was
			used to train the layers.

			Perhaps closest in interpretation to the GRBM as well as to the DBN
			is the mixture of isotropic Gaussian distributions (MoIG) with
			identical covariance and varying mean. Note that after the
			parameters of the GRBM have been fixed, adding layers to the DBN
			only affects the prior distribution over the means learned by the
			GRBM, but has no effect on their positions. As for the GRBM, the
			scale parameter common to all Gaussian mixture components was chosen
			by cross-validation. Other models taken into account are mixtures of
			Gaussians with unconstrained covariance but zero mean (MoG), and
			mixtures of elliptically contoured Gamma distributions with zero
			mean (MoEC) \citep{Hosseini:2007}.

			The results in Figure \ref{figcomparison} suggest that mixture
			components with freely varying covariance are better suited for
			capturing the structure of 4x4 image patches than mixture components
			with fixed covariance. Strikingly, the DBN with 100 hidden units in
			each layer yielded an even larger log-loss than the MoG-2 model. On
			the other hand, both the DBN and the GRBM outperform the MoIG-100
			model, which in contrast to MoG-2 adjusted the means but not the
			covariance.

			Due to the need to estimate the SRBM's marginals, the estimate of
			the DBN's performance might still be too optimistic.  As the
			right plot in \figref{figis} indicates, the true log-loss is likely
			to be a bit larger. Also note that by using more hidden units, the
			performance of both the GRBM and the DBN might still improve. Of
			course, the same is true for the mixture models, whose performance
			might also be improved by taking more components.

			Without lateral connections, that is, with RBMs instead of SRBMs,
			adding layers to the network only decreased the overall performance.
			For a model with 100 hidden units in each layer, trained with CD(1)
			and the same learning parameters as for the model with lateral
			connections, we estimated the average log-loss to be approximately
			$1.945 \pm 4.3\text{E-3}$ (mean $\pm$ SEM, averaged over 10 trials).
			This suggests that the lateral connections did indeed help to
			improve the performance of the model.

		\subsection{Effect of Additional Layers}
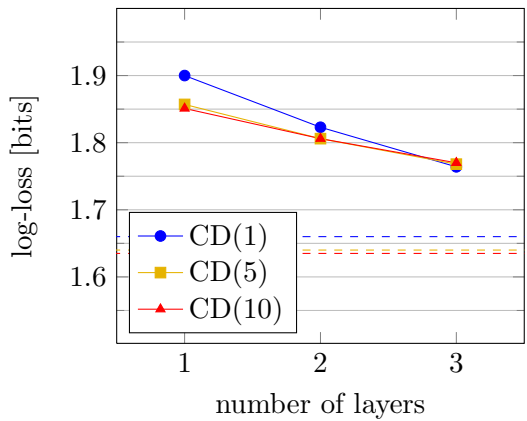
\begin{figure}[t]
	\centering
	\hspace{-1.5cm}
	\begin{tikzpicture}
		\begin{axis}[
				width=7cm,
				ylabel={log-loss [bits]},
				xlabel={number of layers},
				xtick={1,2,3},
				ytick={1.5,1.6,1.7,1.8,1.9},
				minor y tick num=1,
				ymajorgrids=true,
				yminorgrids=true,
				xmin=0.5,
				xmax=3.5,
				ymin=1.501,
				ymax=2.0,
				legend style={
					at={(0.03, 0.03)},
					anchor=south west,
					cells={anchor=west}
				},
				cycle list={
					{line1, solid, mark=*},
					{line2, solid, mark=square*},
					{line3, solid, mark=triangle*},
					{line1, dashed},
					{line2, dashed},
					{line3, dashed}
				}
			]
			\addplot coordinates {
				(1, 1.900)
				(2, 1.823)
				(3, 1.764)
			};
			\addlegendentry{CD(1)};

			\addplot coordinates {
				(1, 1.857)
				(2, 1.806)
				(3, 1.768)
			};
			\addlegendentry{CD(5)};

			\addplot coordinates {
				(1, 1.851)
				(2, 1.806)
				(3, 1.770)
			};
			\addlegendentry{CD(10)};

			\addplot coordinates {
				(0, 1.66)
				(4, 1.66)
			};
			\addplot coordinates {
				(0, 1.64)
				(4, 1.64)
			};
			\addplot coordinates {
				(0, 1.635)
				(4, 1.635)
			};
		\end{axis}
	\end{tikzpicture}
	\caption{Estimated performance of three DBN-100 models trained with different learning rules. The
	improvement per layer decreases as each layer is trained more thoroughly. For each learning rule,
	out of 10 trials, only the trial with the best performance is shown. The dashed lines indicate
	the estimated potential log-loss of the first-layer GRBM.}
	\label{figcdlayers}
\end{figure}
			Using better approximations to ML learning by taking larger CD
			parameters led to an improved performance of the GRBM. However, the
			same could not be observed for the three-layer DBN, whose estimated
			performance was almost the same for all tested CD parameters
			(\figref{figcdlayers}).  In other words, adding layers to the
			network was less effective if each layer was trained more
			thoroughly.

			In many cases, adding a third layer led to an even worse performance
			if the model was trained with CD(5) or CD(10). A likely cause for
			this behavior are too large learning rates, leading to a divergence
			of the training process. In Figure \ref{figcdlayers}, only the best
			results are shown, for which the training converged.

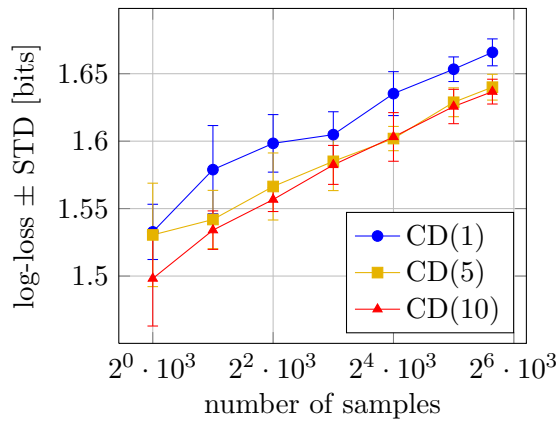
\begin{figure}[t]
	\centering
	\hspace{-1.4cm}
	\begin{tikzpicture}
		\begin{semilogxaxis}[
				width=7cm,
				xlabel={number of samples},
				ylabel={log-loss $\pm$ STD [bits]},
				xtick={1000,4000,16000,64000},
				xticklabels={$2^0 \cdot 10^3$,$2^2 \cdot 10^3$, $2^4 \cdot 10^3$, $2^6 \cdot 10^3$},
				ytick={1.45,1.5,1.55,1.6,1.65},
				xmajorgrids=true,
				ymajorgrids=true,
				ymin=1.4501,
				legend style={
					at={(0.97, 0.03)},
					anchor=south east,
					cells={anchor=west}
				},
				cycle list={
					{line1, solid, mark=*},
					{line2, solid, mark=square*},
					{line3, solid, mark=triangle*}
				}
			]

			\addplot plot[error bars/.cd,y dir=both,y explicit] coordinates {
				(1000, 1.5327462494) +- (0, 0.0204746008479)
				(2000, 1.57888526535) +- (0, 0.0326955668324)
				(4000, 1.59837075841) +- (0, 0.021330204313)
				(8000, 1.60484590404) +- (0, 0.0169622034592)
				(16000, 1.6352493739) +- (0, 0.0162832416624)
				(32000, 1.65333327936) +- (0, 0.0091252306673)
				(50000, 1.66584220253) +- (0, 0.00997484008709)
			};
			\addlegendentry{CD(1)}

			\addplot plot[error bars/.cd,y dir=both,y explicit] coordinates {
				(1000, 1.53050677688) +- (0, 0.0383456187078)
				(2000, 1.5419165797) +- (0, 0.0216105326855)
				(4000, 1.56640546927) +- (0, 0.0248532674108)
				(8000, 1.58505862105) +- (0, 0.0217065290385)
				(16000, 1.60192795681) +- (0, 0.00899988086464)
				(32000, 1.62883868024) +- (0, 0.0106946334409)
				(50000, 1.64002162814) +- (0, 0.00957225281328)
			};
			\addlegendentry{CD(5)}

			\addplot plot[error bars/.cd,y dir=both,y explicit] coordinates {
				(1000, 1.49816953348) +- (0, 0.0352164769585)
				(2000, 1.53403657088) +- (0, 0.0142686930496)
				(4000, 1.55671264541) +- (0, 0.00888070502269)
				(8000, 1.58239994992) +- (0, 0.0144446082214)
				(16000, 1.60313945463) +- (0, 0.0180300451271)
				(32000, 1.6256971912) +- (0, 0.0127365094474)
				(50000, 1.6367264599) +- (0, 0.00918918665128)
			};
			\addlegendentry{CD(10)}
		\end{semilogxaxis}
	\end{tikzpicture}
	\caption{Estimated potential log-loss. Each graph represents the estimated potential log-loss
	of one GRBM, averaged over 10 estimates with different test sets. The size of the data sets used in the
	estimates is given on the horizontal axis. Error bars indicate one standard deviation. After 50000
	samples, the estimates of the potential log-loss have still not converged.}
	\label{figpotential}
\end{figure}

			The estimated improvement of the three-layer DBN over the GRBM is
			about 0.1 bit when trained with CD(5) or CD(10). An important
			question to ask is why the improvement per added layer is so small.
			Insight into this question might be gained by evaluating the
			potential log-likelihood of the GRBM, which represents a practical
			limit to the performance that can be achieved by means of greedy
			learning and can in principle be evaluated even before training any
			additional layers. If the potential log-loss of a trained GRBM is
			close to its log-loss, adding layers is a priori unlikely to prove
			useful. However, exact evaluation of the potential log-likelihood is
			intractable, as it involves two nested integrals with respect to the
			data distribution,
			\begin{align*}
				\int \tilde p(x) \log \int  \tilde p(x_0) q_0(x \mid x_0) dx_0\ dx.
			\end{align*}
			Nevertheless, using optimistic estimates, we were still able to
			infer something about the DBN's capability to improve over the GRBM:
			We estimated the potential log-likelihood using the same set of data
			samples to approximate both integrals, thereby encouraging
			optimistic estimation. Note that estimating the potential
			log-likelihood in this manner is similar to evaluating the
			log-likelihood of a kernel density estimate on the training data,
			although the reconstruction distribution $q_0(x \mid x_0)$ might not
			correspond to a valid kernel. Also note that by taking more and more
			data samples, the estimate of the potential log-loss should become
			more and more accurate. Figure \ref{figpotential} indicates that the
			potential log-loss of a GRBM with 100 hidden units and trained with
			CD(1) is at least 1.66 or larger, which is still worse than the
			performance of, for example, the mixture of Gaussian distribution
			with 5 components.

			Ideally, while training the first layer, one would like to take into
			account the fact that more layers will be added to the
			network. The potential log-loss suggests a regularization which
			minimizes the reconstruction error. Given that a model with perfect
			reconstruction is a fixed point of CD learning
			\citep{LeRoux:2008p7356} and considering the fact that a DBN trained
			with CD(1) led to the same performance as a DBN trained with CD(10)
			(\figref{figcdlayers}), one might hope that CD already has such a
			regularizing effect. As the left plot in Figure \ref{figpotential}
			shows, however, this could not be confirmed: Better approximations
			to ML learning led to a better estimated potential log-loss.

	\section{Discussion}
		We have shown how the likelihood of DBNs can be estimated in a way that is
		both tractable and simple enough to be used in practice. Reliable
		estimators for the likelihood are an important tool not only for the
		evaluation of models deployed in density estimation tasks, but also for
		the evaluation of the effect of different training settings and learning
		rules which try to optimize the likelihood. Thus, the introduced estimator
		potentially adds to the toolbox of everyone training DBNs and facilitates
		the search for better learning algorithms by allowing one to evaluate
		their effect on the likelihood directly.

		However, in cases where models with intractable unnormalized marginal
		distributions are used to build up a DBN, estimating the likelihood of
		DBNs with three or more layers is still a difficult problem. More
		efficient ways to estimate the unnormalized marginals will be required if
		the proposed estimator is to be used with much larger models than the ones
		discussed in this article.  In the common case where a DBN is solely based
		on RBMs, this problem does not occur and the estimator is readily
		applicable.

		We have provided evidence that a particular DBN is not very well suited
		for the task of modeling natural image patches if the goal is to do
		density estimation. Furthermore, we have shown that adding layers to the
		network improves the overall performance of the model only by a small
		margin, especially if the lower layers are trained thoroughly. By
		estimating the potential log-loss---a joint property of the trained
		first-layer model and the greedy learning procedure---we showed that even
		with a lower-bound optimal model in the second layer, the overall
		performance of the DBN would have been unlikely to be much better.

		The potential log-loss suggests two possible ways to improve the training
		procedure: On the one hand, the lower layers might be regularized so as to
		keep the potential improvement that can be achieved with greedy learning
		large. On the other hand, the lower bound optimized during greedy learning
		might be replaced with a different objective function which represents a
		better approximation to the true likelihood. Future research will have to
		show whether these approaches are feasible and can lead to measurable
		improvements.

		The research on hierarchical models of natural images is still in its
		infancy. Although several other attempts have been made to create
		multi-layer models of natural images
		\citep{Sinz:2010p7264,Koster:2010p7346,Hinton:2006p6649,Karklin:2005p7429},
		these models have either been (by design) limited to two layers, or a
		substantial improvement beyond two layers has not been found. Instead, the
		optimization and creation of new shallow architectures has so far proven
		more fruitful. It remains to be seen whether this apparent limitation of
		hierarchical models will be overcome by, for example, creating models and
		more efficient learning procedures that can be used with larger patch
		sizes, or whether this observation is due to a more fundamental problem
		related to the task of estimating the density of natural images.

	\newpage
	\acks{This work is supported by the German Ministry of Education, Science, Research and Technology through
	the Bernstein award to Matthias Bethge (BMBF, FKZ: 01GQ0601) and the Max Planck Society.}

	\appendix
	\section*{Appendix A.}
	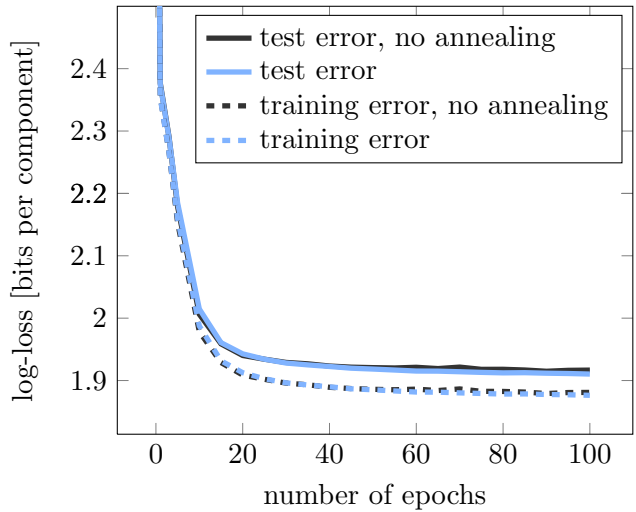
\begin{figure}[h]
		\centering
		\begin{tikzpicture}
			\begin{axis}[
					xlabel={number of epochs},
					ylabel={log-loss [bits per component]},
					ymax=2.5,
					ytick={1.9,2.0,2.1,2.2,2.2,2.3,2.4},
					legend style={
						cells={anchor=west}
					}
				]

				\addplot plot [color=line5, no markers, line width=2] coordinates {
					(0, 3.24019098232) +- (0, 0.000417756971942)
					(1, 2.37668341248) +- (0, 0.0141048115649)
					(3, 2.2913619003) +- (0, 0.0153142477835)
					(5, 2.17872036686) +- (0, 0.0125902334944)
					(10, 2.00620464895) +- (0, 0.00567246224339)
					(15, 1.95878104041) +- (0, 0.0043332102259)
					(20, 1.94017803598) +- (0, 0.00380753853929)
					(25, 1.93418892697) +- (0, 0.00593790518635)
					(30, 1.92900472317) +- (0, 0.00564522599307)
					(35, 1.92678023073) +- (0, 0.0075055346686)
					(40, 1.92323413567) +- (0, 0.00631376685841)
					(45, 1.92143864395) +- (0, 0.00449889655434)
					(50, 1.9206780627) +- (0, 0.00538782773138)
					(55, 1.92013923729) +- (0, 0.00767272702541)
					(60, 1.9212288781) +- (0, 0.00766713632191)
					(65, 1.91915854322) +- (0, 0.00774774435673)
					(70, 1.92153529891) +- (0, 0.00999991191487)
					(75, 1.91793267945) +- (0, 0.00777101325728)
					(80, 1.91802152515) +- (0, 0.00599199630524)
					(85, 1.91687090503) +- (0, 0.0079108599631)
					(90, 1.91486495053) +- (0, 0.00567283101142)
					(95, 1.91635195503) +- (0, 0.0066039238443)
					(100, 1.916711459) +- (0, 0.00559549572104)
				};
				\addlegendentry{test error, no annealing};

				\addplot plot [color=line6, no markers, line width=2] coordinates {
					(0, 3.24030346626) +- (0, 0.000449970799256)
					(1, 2.3747138971) +- (0, 0.0156111444583)
					(3, 2.29007586067) +- (0, 0.0123361500932)
					(5, 2.18467182718) +- (0, 0.0170551409542)
					(10, 2.01467352396) +- (0, 0.00706127657507)
					(15, 1.96072897659) +- (0, 0.00368215883269)
					(20, 1.94277372382) +- (0, 0.00556117003031)
					(25, 1.93422895177) +- (0, 0.00611650054361)
					(30, 1.92804996927) +- (0, 0.00511258955627)
					(35, 1.92509871222) +- (0, 0.00369284127679)
					(40, 1.92253167679) +- (0, 0.00427932415526)
					(45, 1.91977829623) +- (0, 0.00577403963795)
					(50, 1.91840514091) +- (0, 0.00376840363593)
					(55, 1.91683859062) +- (0, 0.00272563567603)
					(60, 1.91520990602) +- (0, 0.00333812446771)
					(65, 1.91506180956) +- (0, 0.00532784874774)
					(70, 1.91410346994) +- (0, 0.00322735594813)
					(75, 1.91308379359) +- (0, 0.00234452646631)
					(80, 1.9122890999) +- (0, 0.00321258869226)
					(85, 1.91258233291) +- (0, 0.00389948276748)
					(90, 1.9117742306) +- (0, 0.00248569662208)
					(95, 1.91125706443) +- (0, 0.00277214303638)
					(100, 1.91039559153) +- (0, 0.00282491902503)
				};
				\addlegendentry{test error};

				\addplot plot [color=line5, dashed, no markers, line width=2] coordinates {
					(0, 3.20829397022) +- (0, 0.000464561830591)
					(1, 2.35540035896) +- (0, 0.0146696806326)
					(3, 2.26708768679) +- (0, 0.0156229488593)
					(5, 2.15309780742) +- (0, 0.0120568052669)
					(10, 1.97840703427) +- (0, 0.00505162134804)
					(15, 1.92882894564) +- (0, 0.00488705626769)
					(20, 1.90880576806) +- (0, 0.00439230162711)
					(25, 1.9019788235) +- (0, 0.00596794841614)
					(30, 1.89586071635) +- (0, 0.00521526379937)
					(35, 1.893178233) +- (0, 0.00741576157417)
					(40, 1.88917302316) +- (0, 0.00625856016951)
					(45, 1.8870928966) +- (0, 0.00394252439336)
					(50, 1.88592992197) +- (0, 0.00499988008771)
					(55, 1.88521087404) +- (0, 0.00651409865972)
					(60, 1.88617462864) +- (0, 0.00704970605018)
					(65, 1.88396959448) +- (0, 0.00766464303288)
					(70, 1.88652541232) +- (0, 0.0105695548081)
					(75, 1.88271197466) +- (0, 0.00735148450217)
					(80, 1.88252073838) +- (0, 0.00515756308174)
					(85, 1.88149839839) +- (0, 0.00829636852471)
					(90, 1.87914492496) +- (0, 0.00543699635706)
					(95, 1.88075172742) +- (0, 0.00595154136132)
					(100, 1.8809503827) +- (0, 0.00591513018793)
				};
				\addlegendentry{training error, no annealing};

				\addplot plot [color=line6, dashed, no markers, line width=2] coordinates {
					(0, 3.20840695831) +- (0, 0.000441427317018)
					(1, 2.35299631994) +- (0, 0.0151449541822)
					(3, 2.26613100212) +- (0, 0.012617752588)
					(5, 2.15946968138) +- (0, 0.0173927880704)
					(10, 1.98755161949) +- (0, 0.00766404286735)
					(15, 1.93163022485) +- (0, 0.00330888957314)
					(20, 1.91256990609) +- (0, 0.00476752689804)
					(25, 1.90281334777) +- (0, 0.00615313971047)
					(30, 1.8962546748) +- (0, 0.00586250951046)
					(35, 1.89286206266) +- (0, 0.00286624701287)
					(40, 1.88992373402) +- (0, 0.00459968486499)
					(45, 1.88677947492) +- (0, 0.00519985133503)
					(50, 1.88518609688) +- (0, 0.00318020361785)
					(55, 1.88330847362) +- (0, 0.00272217856954)
					(60, 1.88154833118) +- (0, 0.0028373414513)
					(65, 1.8813520979) +- (0, 0.00524681788854)
					(70, 1.88020254578) +- (0, 0.00278490765974)
					(75, 1.87925458854) +- (0, 0.0022308386073)
					(80, 1.87830608799) +- (0, 0.00306613358865)
					(85, 1.87854998746) +- (0, 0.00302739362402)
					(90, 1.87763797903) +- (0, 0.00307265089015)
					(95, 1.87719156441) +- (0, 0.00205851700141)
					(100, 1.87629871445) +- (0, 0.00309271745062)
				};
				\addlegendentry{training error};
			\end{axis}
		\end{tikzpicture}
		\caption{Log-loss of a GRBM-100 versus number of epochs, averaged over 10 trials
		using the same training and test sets in each trial. After 50 epochs, the log-loss has
		largely converged. No overfitting could be observed. Using a constant learning rate instead
		of a linearly decreasing learning rate had no effect on the convergence, which means that the
		convergence is not just due to the annealing.}
		\label{figepochs}
	\end{figure}
		In the following, we will summarize the relevant learning as well as
		evaluation parameters used in the experiments of the experimental section.

		The layers of the deep belief network with 100 hidden units were trained
		for 100 epochs. The learning rates were decreased from $1 \cdot 10^{-2}$
		to $1 \cdot 10^{-4}$ during training using a linear annealing schedule.
		As can be seen in Figure \ref{figepochs}, the performance of the GRBM
		largely converged after 50 epochs.

		The covariance of the conditional distribution of the GRBM's visible units
		given the hidden units was fixed to $\sigma I$. $\sigma$ was treated as a
		hyperparameter and chosen via cross-validation with respect to the
		likelihood of the GRBM after all other parameters had been fixed. Weight
		decay of $0.01$ times the learning rate was applied to all weights, but
		not to the biases, and a momentum factor of $0.9$ was used for all
		parameters.  The biases of the hidden units of all layers were initialized
		to be $-1$ as a (rough) means to encourage sparseness.

		As described in Section \ref{secexp}, the second-layer SRBM was
		initialized so that its marginal distribution over the units it shares
		with the GRBM is the same as the marginal distribution defined by the
		GRBM. During training, approximate samples from the visible conditional
		distribution of the SRBMs were obtained using 20 parallel mean field
		updates with a damping parameter of 0.2 \citep{Welling:2002p7082}. During
		evaluation, sequential Gibbs updates were used.

		For the evaluation of the partition function and the marginals, we used
		AIS. The number of intermediate annealing distributions was 1000 in each
		layer. We used a linear annealing schedule, that is, the annealing weights
		determining the intermediate distributions were equally spaced. Though
		this schedule is not optimal from a theoretical perspective
		\citep{Neal:2001p7298}, we only found a small effect on the estimator's
		performance by taking different schedules. The number of AIS samples used
		during the experiments was 100 for the GRBM, 1000 for the third-layer SRBM
		and 100000 for the second-layer SRBM. The number of second-layer AIS
		samples had to be much larger because the samples were used not only to
		estimate the partition function, but also to estimate the second-layer
		SRBM's hidden marginals. As can be inferred from Figure \ref{figis}, even
		after taking this many samples the estimates of the three-layer DBN's
		performance were still somewhat optimistic.

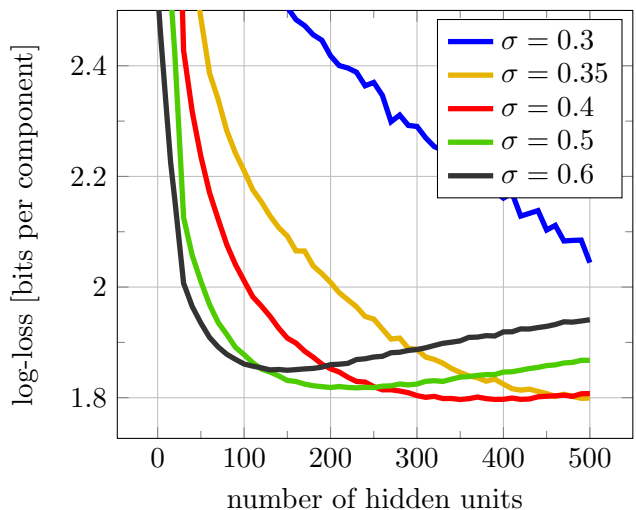
\begin{figure}[t]
	\centering
	\begin{tikzpicture}
		\begin{axis}[
				xlabel={number of hidden units},
				ylabel={log-loss [bits per component]},
				ymax=2.5,
				minor y tick num=1,
				minor x tick num=1,
				ymajorgrids=true,
				xmajorgrids=true,
				legend style={
					cells={anchor=west}
				}
			]

			\addplot plot [color=line1, no markers, line width=2] coordinates {
				(1, 7.16287484452)
				(3, 6.77749204662)
				(5, 6.42436119822)
				(10, 5.54719933027)
				(15, 4.74851822932)
				(20, 4.48888947793)
				(30, 3.26808270968)
				(40, 3.07675734282)
				(50, 2.93173036664)
				(60, 2.83943206301)
				(70, 2.75395081711)
				(80, 2.70143078371)
				(90, 2.66227922579)
				(100, 2.60708305866)
				(110, 2.59512082578)
				(120, 2.55149509607)
				(130, 2.53137114055)
				(140, 2.51233001618)
				(150, 2.50548969029)
				(160, 2.48342892735)
				(170, 2.47256141105)
				(180, 2.45599787517)
				(190, 2.44609436642)
				(200, 2.4179976677)
				(210, 2.40037363346)
				(220, 2.39596188869)
				(230, 2.38836233052)
				(240, 2.36381529462)
				(250, 2.36991647958)
				(260, 2.34636930122)
				(270, 2.29880908746)
				(280, 2.31094240628)
				(290, 2.29210331233)
				(300, 2.29038137913)
				(310, 2.26987807441)
				(320, 2.25392554909)
				(330, 2.24707161893)
				(340, 2.23968140953)
				(350, 2.21002705679)
				(360, 2.2112991379)
				(370, 2.20227007534)
				(380, 2.1807452084)
				(390, 2.17376141136)
				(400, 2.16076039201)
				(410, 2.17001427921)
				(420, 2.12830680282)
				(430, 2.13320560909)
				(440, 2.13818495512)
				(450, 2.10309929373)
				(460, 2.11148177615)
				(470, 2.08330037719)
				(480, 2.08433773829)
				(490, 2.0848812894)
				(500, 2.04405958465)
			};
			\addlegendentry{$\sigma = 0.3$}

			\addplot plot [color=line2, no markers, line width=2] coordinates {
				(1, 5.38872615809)
				(3, 5.13317968292)
				(5, 4.88722046071)
				(10, 4.31429212566)
				(15, 3.78654230882)
				(20, 3.57952568434)
				(30, 2.73035070176)
				(40, 2.58892678581)
				(50, 2.48601377574)
				(60, 2.38654723116)
				(70, 2.33919734391)
				(80, 2.28241123307)
				(90, 2.24215144819)
				(100, 2.21011615976)
				(110, 2.17579876308)
				(120, 2.15414318191)
				(130, 2.12866885882)
				(140, 2.10758769966)
				(150, 2.09217895411)
				(160, 2.06540261088)
				(170, 2.06513252833)
				(180, 2.03801778299)
				(190, 2.02423413577)
				(200, 2.00841872167)
				(210, 1.98981747148)
				(220, 1.97707793068)
				(230, 1.96432912015)
				(240, 1.94691826424)
				(250, 1.94217473689)
				(260, 1.92405044793)
				(270, 1.90605212272)
				(280, 1.90777514679)
				(290, 1.88912627384)
				(300, 1.88757579856)
				(310, 1.8735516387)
				(320, 1.86918475969)
				(330, 1.86243201521)
				(340, 1.85273080794)
				(350, 1.84567995777)
				(360, 1.8396304997)
				(370, 1.83611726794)
				(380, 1.82984640865)
				(390, 1.83368589242)
				(400, 1.82450706714)
				(410, 1.81631136706)
				(420, 1.81366274211)
				(430, 1.81545964288)
				(440, 1.81146247303)
				(450, 1.80747723468)
				(460, 1.80234627437)
				(470, 1.80628348938)
				(480, 1.80246889323)
				(490, 1.79887996052)
				(500, 1.79949257144)
			};
			\addlegendentry{$\sigma = 0.35$}

			\addplot plot [color=line3, no markers, line width=2] coordinates {
				(1, 4.284469438)
				(3, 4.10898327235)
				(5, 3.9423999929)
				(10, 3.55180958492)
				(15, 3.18874212128)
				(20, 2.97119815527)
				(30, 2.42755458731)
				(40, 2.31683462746)
				(50, 2.23608015319)
				(60, 2.170776238)
				(70, 2.12262772885)
				(80, 2.07646259033)
				(90, 2.04068656392)
				(100, 2.01128462105)
				(110, 1.98246943424)
				(120, 1.96568263595)
				(130, 1.94713253433)
				(140, 1.92572991116)
				(150, 1.90775194722)
				(160, 1.89879447939)
				(170, 1.88423130566)
				(180, 1.87315662706)
				(190, 1.86074679234)
				(200, 1.85238273337)
				(210, 1.84642839167)
				(220, 1.83701935054)
				(230, 1.8294278201)
				(240, 1.82829592629)
				(250, 1.82002874473)
				(260, 1.81424025181)
				(270, 1.81440150093)
				(280, 1.81214394766)
				(290, 1.80902516492)
				(300, 1.80384586941)
				(310, 1.80071931195)
				(320, 1.80244494319)
				(330, 1.79850996064)
				(340, 1.79859071096)
				(350, 1.79682413327)
				(360, 1.79826396003)
				(370, 1.80126327345)
				(380, 1.79837875191)
				(390, 1.79690632662)
				(400, 1.79717110902)
				(410, 1.79936463709)
				(420, 1.79704276041)
				(430, 1.79771496971)
				(440, 1.80164738964)
				(450, 1.80269796421)
				(460, 1.80344108881)
				(470, 1.80452982278)
				(480, 1.80245878031)
				(490, 1.80713280204)
				(500, 1.80754809001)
			};
			\addlegendentry{$\sigma = 0.4$}

			\addplot plot [color=line4, no markers, line width=2] coordinates {
				(1, 3.08012707699)
				(3, 2.99345862815)
				(5, 2.91409035341)
				(10, 2.7183375975)
				(15, 2.53328172014)
				(20, 2.41837652922)
				(30, 2.12567146025)
				(40, 2.05722179533)
				(50, 2.01028360692)
				(60, 1.96830612708)
				(70, 1.93592529534)
				(80, 1.91431558552)
				(90, 1.89102699223)
				(100, 1.87782097755)
				(110, 1.8633637813)
				(120, 1.85184833946)
				(130, 1.84634795196)
				(140, 1.83946151332)
				(150, 1.83118701808)
				(160, 1.82937414611)
				(170, 1.82460159975)
				(180, 1.82163728964)
				(190, 1.82003478674)
				(200, 1.81843044818)
				(210, 1.820222063)
				(220, 1.81854890805)
				(230, 1.81782966231)
				(240, 1.81879205366)
				(250, 1.81830130467)
				(260, 1.82089998957)
				(270, 1.82144145021)
				(280, 1.82488994148)
				(290, 1.82319070965)
				(300, 1.82476624943)
				(310, 1.82960645637)
				(320, 1.83073905583)
				(330, 1.82944288778)
				(340, 1.83359067959)
				(350, 1.83722872417)
				(360, 1.83826211836)
				(370, 1.84034043653)
				(380, 1.84178032518)
				(390, 1.84182500953)
				(400, 1.84634349403)
				(410, 1.84698632886)
				(420, 1.84969019701)
				(430, 1.85277918892)
				(440, 1.85479258176)
				(450, 1.85811719806)
				(460, 1.85931396728)
				(470, 1.86281335536)
				(480, 1.86388596448)
				(490, 1.86757232654)
				(500, 1.86765521283)
			};
			\addlegendentry{$\sigma = 0.5$}

			\addplot plot [color=line5, no markers, line width=2] coordinates {
				(1, 2.5096814425)
				(3, 2.46476789903)
				(5, 2.42418838484)
				(10, 2.32352731076)
				(15, 2.22538534066)
				(20, 2.15561390512)
				(30, 2.00696725641)
				(40, 1.96480438066)
				(50, 1.93492855613)
				(60, 1.90902242358)
				(70, 1.89177287183)
				(80, 1.87848804433)
				(90, 1.86933622072)
				(100, 1.86090003052)
				(110, 1.85710852491)
				(120, 1.8540425636)
				(130, 1.85030547002)
				(140, 1.85101251337)
				(150, 1.84966315083)
				(160, 1.85079749148)
				(170, 1.85207976367)
				(180, 1.85303144496)
				(190, 1.85525879542)
				(200, 1.85990972662)
				(210, 1.86054998991)
				(220, 1.86208496011)
				(230, 1.86898387371)
				(240, 1.87040574351)
				(250, 1.87362053334)
				(260, 1.87481854193)
				(270, 1.88137838071)
				(280, 1.88220605701)
				(290, 1.88545367719)
				(300, 1.8871174962)
				(310, 1.89021811325)
				(320, 1.89437943833)
				(330, 1.89838659138)
				(340, 1.90039276578)
				(350, 1.90277921973)
				(360, 1.90903978497)
				(370, 1.9095873795)
				(380, 1.91218796502)
				(390, 1.91159502921)
				(400, 1.91901552496)
				(410, 1.91919048949)
				(420, 1.92440749911)
				(430, 1.92420819109)
				(440, 1.92680957772)
				(450, 1.92899795211)
				(460, 1.93232572413)
				(470, 1.9371861539)
				(480, 1.93663027003)
				(490, 1.93867325215)
				(500, 1.94098366435)
			};
			\addlegendentry{$\sigma = 0.6$}
		\end{axis}
	\end{tikzpicture}
	\caption{Joint evaluation of the number of hidden units and the component variance. By taking more hidden units
	and smaller variances, the performance of the GRBM can still be improved. All models were trained for 50 epochs using CD(1).}
	\label{figcrv}
\end{figure}
		Lastly, note that the performance of the GRBM and the DBN might still be
		improved by taking a larger number of hidden units. A post-hoc analysis
		revealed that the GRBM does indeed not overfit but continues to improve
		its performance if the variance is decreased while increasing the number
		of hidden units (\figref{figcrv}).

		Code for training and evaluating deep belief networks using the estimator
		presented in this article can be found under
		\begin{center}
			\url{http://kyb.tuebingen.mpg.de/bethge/code/dbn/dbn.tar.gz}.
		\end{center}

		\newpage

	\vskip 0.2in
	\bibliography{references}
\end{document}